\documentclass{bmvc2k}

\usepackage{floatrow}
\usepackage{graphicx}
\usepackage{amsmath}
\usepackage{amssymb}
\usepackage{booktabs}
\usepackage{multirow}
\usepackage{caption}
\usepackage{pifont}% http://ctan.org/pkg/pifont
\usepackage{wrapfig}
\usepackage[ruled]{algorithm2e}

\title{Learning to Augment via Implicit Differentiation for Domain Generalization}

\addauthor{Tingwei Wang}{tingwei.wang@surrey.ac.uk}{1}
\addauthor{Da Li}{dali.academic@gmail.com}{2}
\addauthor{Kaiyang Zhou}{k.zhou.vision@gmail.com}{3}
\addauthor{Tao Xiang}{t.xiang@surrey.ac.uk}{1}
\addauthor{Yi-Zhe Song}{y.song@surrey.ac.uk}{1}

\addinstitution{
 University of Surrey\\
 Guildford, UK
}
\addinstitution{
 Samsung AI Center\\
 Cambridge, UK
}
\addinstitution{
 Nanyang Technological University\\
 Singapore
}

\runninghead{WANG, LI, ZHOU, XIANG, SONG}{LEARNING TO AUGMENT}

\begin{document}

\maketitle

\begin{abstract}
Machine learning models are intrinsically vulnerable to domain shift between training and testing data, resulting in poor performance in novel domains. Domain generalization (DG) aims to overcome the problem by leveraging multiple source domains to learn a domain-generalizable model. In this paper, we propose a novel augmentation-based DG approach, dubbed AugLearn. Different from existing data augmentation methods, our AugLearn views a data augmentation module as hyper-parameters of a classification model and optimizes the module together with the model via meta-learning. Specifically, at each training step, AugLearn (i) divides source domains into a pseudo source and a pseudo target set, and (ii) trains the augmentation module in such a way that the augmented (synthetic) images can make the model generalize well on the pseudo target set. Moreover, to overcome the expensive second-order gradient computation during meta-learning, we formulate an efficient joint training algorithm, for both the augmentation module and the classification model, based on the implicit function theorem. With the flexibility of augmenting data in both time and frequency spaces, AugLearn shows effectiveness on three standard DG benchmarks, PACS, Office-Home and Digit-DG.
\end{abstract}

%-------------------------------------------------------------------------
\section{Introduction}
\label{sec:intro}
Humans excel at learning visual concepts that are generalizable across different scenarios and environments. For instance, we can easily recognize a dog image no matter whether the image is a realistic photo, a cartoon or even a human drawn sketch. However, though deep neural networks have achieved great success in many computer vision tasks, their ability to generalize to novel data distributions remains rather limited. This hampers the wide deployment of deep learning models in real-world applications. The root of poor generalization is the domain shift problem~\cite{taori2020measuring, ben2010theory, moreno2012unifying, recht2019imagenet} which is known by machine learning researchers for decades. One solution to the domain shift problem is unsupervised domain adaptation (UDA)~\cite{ganin2015unsupervised, gong2012geodesic, long2014transfer, baktashmotlagh2013unsupervised}, which exploits unlabeled target domain data for domain adaptation. Although UDA avoids target domain data annotation, it still needs access to the target domain data to perform model adaptation for each target domain. Domain generalization (DG)~\cite{blanchard2011generalizing, zhou2021survey, matsuura2020domain, huang2020self, ding2017deep, jeon2021feature} is motivated to solve this drawback of UDA. Given multiple source domains, the goal of DG is to learn a model that can generalize well to any unseen target domain without any model adaptation. 

Most existing DG methods are either feature alignment or meta-learning based. Feature alignment based methods borrow ideas from the domain adaptation community to align features across source domains~\cite{motiian2017unified,ghifary2016scatter,li2018adversarial, erfani2016robust, jin2020feature, otalora2019staining}. These methods focus on minimizing the divergence between the source domains in feature space, which ensures that the extracted representations are domain agnostic. In contrast, meta-learning methods expose models to domain shift during training~\cite{balaji2018metareg, dou2019domain, li2019episodic, li2019feature}. The underlying idea is to split training data into meta-train and meta-test sets without overlapping domains. A DG model is trained on the meta-train set in a way such that its loss on the meta-test set is also low. 

Recently, data augmentation based DG methods have attracted increasing attention~\cite{zakharov2019deceptionnet,zhou2020learning,zhou2021domain, volpi2019addressing, zhang2020generalizing, yue2019domain}. Existing methods design image synthesizers to synthesize images that do not belong to any existing source domains. They are designed with the assumption that diversified source domains enable the model to learn more generalizable features. 
However, existing augmentation DG methods normally require some complicated design of learning objectives, such as differing the augmented images from the source data, maintaining the fidelity of the augmented images, retaining the same semantic meaning of the augmented image to the vanilla image. More importantly, there is no guarantee that the augmented images, when used for training the main classification model, can ensure the model generalizes well to an unseen domain.

In this paper, we propose a novel meta learning based augmentation method for DG. Different from the existing augmentation based DG methods, we treat the image augmentation module as the hyperparameters of a classification model and optimize both jointly. Crucially, the augmentation module is optimized explicitly to help the model generalize to a novel domain. To guarantee generalization, we expose the model to simulated domain shift and meta learn the augmentation module to minimize the generalization error of the model. 
However, optimizing the hyperparameters of the model is non-trivial when these hyperparameters are actually parameters of a deep CNN (i.e., the augmentation module) and thus in the order of millions. In particular, the typical bilevel hyperparameter optimization will produce second-order gradients which poses a serious computational challenge \cite{luketina2016scalable, rajeswaran2019meta, lorraine2020optimizing}. To address this computational issue, we employ the implicit function theorem (IFT) \cite{luketina2016scalable} to avoid storing the inner loop update trajectories which is prohibitively expensive. Specifically, we create episodes from minibatch images of source domains during training. In each episode, we randomly split the images from different source domains into pseudo source and pseudo target domains. The classification model is optimized by minimizing the classification loss on pseudo source domain data. The hyperparameters (augmentation module) are then optimized using the Neumann series approximated IFT by minimizing a validation loss on the pseudo target data. Frequency-based data augmentation has recently shown promising performance in domain generalization~\cite{xu2021fourier}. Flexibly, our AugLearn is not only applicable to augment the input image in time space, but also in frequency space. We can simply feed the frequency spectrum, such as obtained by discrete cosine transformation (DCT), of the input image into the AugLearn and then inverse the augmented frequency spectrum to the time space -- this variant is dubbed AugLearn-F. Our proposed augmentation module is model-agnostic and can be applied to any base DG methods.

Our {\bf contributions} are summarized as follows: (1) We propose a novel DG framework in which the augmentation module is viewed as hyperparameters of the model and optimized efficiently using the implicit function theorem (IFT). (2) Different from the most conventional data augmentation methods, focusing on the augmentation in the time space only, our AugLearn is capable of augmenting the input data in both the time and frequency spaces.
(3) Our proposed module is model-agnostic and applicable to any base DG methods. Extensive experiments are carried out, and the results show that our method achieves the state of the art performance on two popular DG benchmarks, PACS and Digits-DG.

\section{Related Work}
\paragraph{Domain generalization.}
The DG problem was first introduced in ~\cite{blanchard2011generalizing} in which they proposed a kernel-based DG approach. Since then DG has been receiving increasing attention~\cite{zhou2021survey, wang2021generalizing} from the research community due to its importance to practical machine learning applications. DG models aim at extracting general representations that can perform well on unseen target domains. In general, the existing DG methods can be categorised into three groups, namely domain alignment, data augmentation and meta-learning.

(1) The alignment based methods are mainly inspired by the domain adaptation literature. The main idea is to learn representations that are domain-agnostic among source domains such that the representations can generalize to any unseen target domain. Muandet \emph{et al.}~\cite{muandet2013domain} developed Domain-Invariant Component Analysis (DICA), which adopts a kernel-based optimization algorithm that reduces the discrepancy across source domains, to learn domain invariant features. Li \emph{et al.}~\cite{li2017deeper} proposed a low-rank parameterized CNN for learning domain agnostic features. Li \emph{et al.}~\cite{li2018adversarial} extended adversarial autoencoders by imposing the Maximum Mean Discrepancy (MMD) measure to align distributions among different source domains. Motiian \emph{et al.}~\cite{motiian2017unified} employed a Siamese architecture to map different source domains to a discriminative embedding subspace, where the mapped features are semantically aligned and maximally separated.

(2) The existing data augmentation based DG methods mainly focused on two folds, the image level and feature level augmentations. Yue \emph{et al.}~\cite{yue2019domain} proposed a domain randomization based augmentation to diversify the input images and forced the model to learn domain invariant features. 
%Zakharov \emph{et al.}~\cite{zakharov2019deceptionnet} designed a min-max optimization method that could select the most destructive augmentations to improve the training model. 
Zhou \emph{et al.}~\cite{zhou2020deep} developed a learnable data perturbation module to generate novel images to improve the model generalization, and a novel domain generator using optimal transport as a followup~\cite{zhou2020learning}. Recently researchers found that feature augmentation is effective in improving model generalization~\cite{zhou2021domain, li2021simple}. MixStyle~\cite{zhou2021domain} assumed the feature statistics represent the domain style information and proposed to interpolate feature representations following Mixup~\cite{zhang2017mixup}. Li et al.~\cite{li2021simple} lately found that a simple feature perturbation using Gaussian noise worked pretty well on improving domain generalization. All these augmentation based methods require explicit objectives, such as generating new domain images while retaining the same semantic meaning. Furthermore, those methods focus on diversifying the source domains and this is not promising to improve the generalization ability of the models. Our proposed method optimizes the augmentation module implicitly to be general to unseen domain.

(3) Meta-learning is widely applied on improving few shot learning~\cite{finn2017model, ravi2016optimization} and has now been exploited to tackle domain shift problems~\cite{li2018learning, balaji2018metareg, dou2019domain}. \cite{li2018learning} proposed to mimic the domain shift during training by splitting the source domains into meta train and meta test such that the training model can learn to learn how to address DG. Balaji \emph{et al.}~\cite{balaji2018metareg} then reformulated the DG problem as meta-learning a generalizable regularizer. Li \emph{et al.}~\cite{li2019episodic} developed an episodic training paradigm to improve DG by manipulating domain specific feature extractors and classifiers. Dou \emph{et al.}~\cite{dou2019domain} designed a meta learning algorithm to improve DG by maintaining the inter-class knowledge consistency across source domains. Du \emph{et al.}~\cite{du2020metanorm} proposed to meta learn a normalization to tackle the statistic changes among source and target domains.

\paragraph{Implicit function theorem.}
Bilevel optimization is typical in hyperparameter optimization~\cite{bengio2000gradient} and meta learning~\cite{finn2017model, rajeswaran2019meta}. However, it normally triggers the second order derivatives during the back propagation. Implicit function theorem (IFT) has been explored to address such computational issue~\cite{luketina2016scalable, rajeswaran2019meta, lorraine2020optimizing}. However, the conventional IFT requires the computation of inverse hessian which is intractable in deep neural networks. Therefore, various approximations have been proposed to mitigate this problem. \cite{luketina2016scalable} proposed to approximate the hessian with the identity matrix, while \cite{rajeswaran2019meta} uses Conjugate Gradient (CG) to approximate the vector-inverse hessian product. More recently, \cite{lorraine2020optimizing} proposed a novel approximation using Neumann series, which is used in our paper.

\section{Methodology}
\paragraph{Problem setup.} In the DG setting, it is normally assumed there are multiple source domains $\mathcal{D} = \{{D}_1,\dots ,{D}_S\}$, where each ${D}_i$ typically consists of many data pairs $\{x, y\}_{i}^N$. Then a model $F_\theta$, i.e. CNN, is trained on the source domains, and then tested on an unseen target domain ${D}_{S+1}$.

\paragraph{ERM baseline.} The most straightforward DG method is to train a model using the empirical risk minimization on the source domain data. The formulation is as follows
\begin{equation}
\label{eq:erm}
\underset{\theta}{\arg\min} \frac{1}{|\mathcal{D}|} \sum_{x, y \sim \mathcal{D}} \ell_{ce}(F_\theta(x), y),
\end{equation}
where $\ell_{ce}$ is the cross entropy loss. After training, $F_\theta$ will be deployed for model inference.

\paragraph{Naive data augmentation.}
Data augmentation as a regularization has shown its effectiveness on improving the model generalization~\cite{krizhevsky2012imagenet}. Typically the augmentation operation is sampled and constructed stochastically from different hand crafted operations, such as \texttt{flipping}, \texttt{rotation}, \texttt{jittering} and \texttt{coloring}, denoted as $\mathcal{A}(\cdot)$. Then the augmented images will be used to train the model, thus Eq.~\ref{eq:erm} becomes
\begin{equation}
\underset{\theta}{\arg\min} \frac{1}{|\mathcal{D}|} \sum_{x, y \sim \mathcal{D}, A\sim \mathcal{A}} \ell_{ce}(F_\theta(A(x)), y).
\end{equation}

\subsection{Data Augmentation Module as Hyperparameters}
\begin{figure*}[htb]
    \centering
    \includegraphics[width=0.8\textwidth]{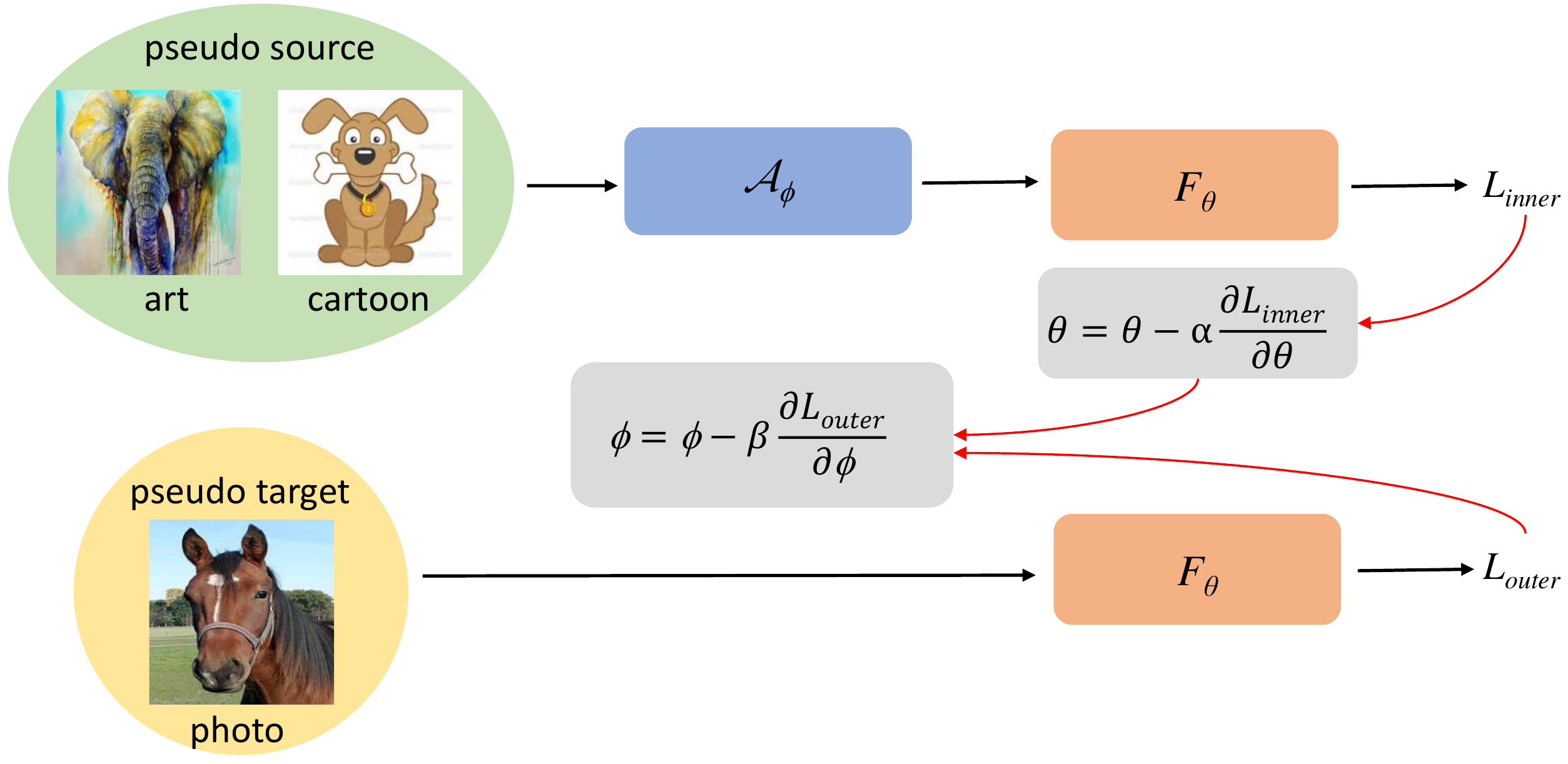}
    \caption{The overview of our proposed AugLearn method. We treat the augmentation module $\mathcal{A}_\phi$ as hyperparameters of the classification model $F_\theta$. The classification model is updated on pseudo source domains, and the augmentation module is optimized on the pseudo target domain under the condition that $\theta^\ast$ is optimal to pseudo source domains. The red arrows in this figure denote the gradient flow through the second-order differentiation.} 
    \label{fig:framework}
\end{figure*}

The typical data augmentation is made up of different hand crafted operations, which have demonstrated their effectiveness in standard supervised learning but may not be optimal to domain generalization tasks. Therefore, we treat the data augmentation module as hyperparameters of our classification model and optimize it against domain shift during training. The overall scheme is illustrated in Figure~\ref{fig:framework}.

\begin{figure*}[h]
    \centering
    \includegraphics[width=0.8\textwidth]{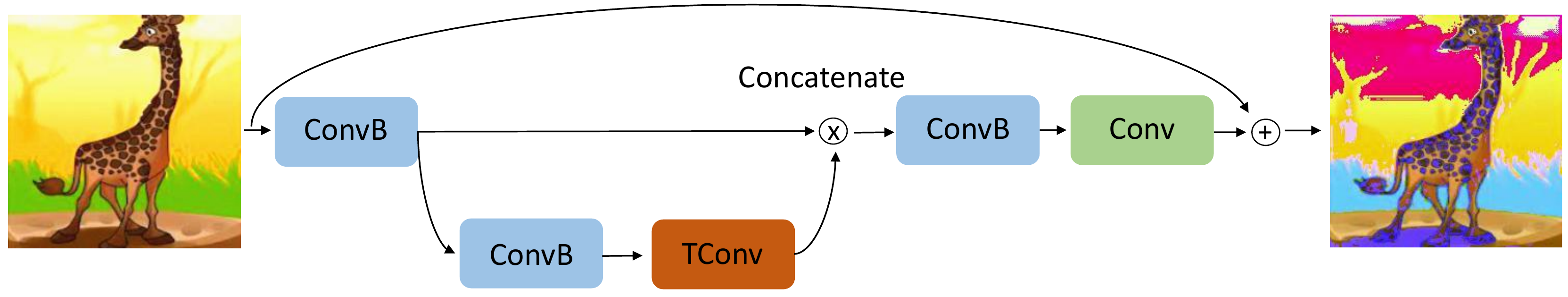}
    \caption{The illustration of the augmentation module. ConvB, TConv and Conv in this figure are convolutional block, transpose convolution layer and convolution layer, respectively.} 
    \label{fig:aug_module}
\end{figure*}

\paragraph{Augmentation module.}
We exploit a simple UNet~\cite{ronneberger2015unet} as our augmentation module, which is demonstrated in Figure~\ref{fig:aug_module}. Specifically, this UNet consists of three convolutional blocks, a transpose convolution layer and a convolution layer. The convolutional block is composed of two convolution layers, each followed by a ReLU activation. In addition, a max-pooling layer is attached to the first convolutional block. The UNet is formulated as $\mathcal{A}_{\phi}(\cdot)$ parameterized by $\phi$.

\paragraph{Bilevel optimization.}
In order to optimize our augmentation module, we create episodes using our source domain data during training. we split the source domain data $\mathcal{D}$ into pseudo source $\hat{\mathcal{D}^{psrc}}$ and pseudo target $\hat{\mathcal{D}^{ptrg}}$ at each mini-batch step. Then the bilevel optimization is conducted, including an inner loop optimization on $\hat{\mathcal{D}^{psrc}}$ which is formulated as 

\begin{equation}
\label{eq:inner}
\theta^*= \underset{\theta}{\arg\min} \frac{1}{|\hat{\mathcal{D}^{psrc}}|} \sum_{x, y \sim \hat{\mathcal{D}^{psrc}}} \ell_{ce}(F_\theta(\mathcal{A}_{\phi}(x)), y),
\end{equation}
and an outer loop optimization on $\hat{\mathcal{D}^{ptrg}}$
\begin{equation}
\label{eq:outer}
\phi= \underset{\phi}{\arg\min} \frac{1}{|\hat{\mathcal{D}^{ptrg}}|} \sum_{x, y \sim \hat{\mathcal{D}^{ptrg}}} \ell_{ce}(F_{\theta^*}({\phi}(x)), y).
\end{equation}
% However, the typical bilevel optimization will produce the second order gradients, which is intractable for deep neural networks, when conducting Eq.~\ref{eq:outer}.

Let us break down the computation in~Eq.\ref{eq:outer}. If the inner and outer loop losses are denoted as $L_{inner}$ and $L_{outer}$, then the hypergradients of $L_{outer}$ w.r.t. the augmentation module $\mathcal{A}_{\phi}$ is computed as
\begin{equation}
\label{eq:outer_gradient}
\begin{aligned}
    \frac{\partial L_{outer}(\phi)}{\partial \phi} & =  \frac{\partial L_{outer}}{\partial \phi}  + \frac{\partial L_{outer}}{\partial \theta}  \frac{\partial \theta^\ast}{\partial{\phi}} \\
    & =\underbrace{\frac{\partial L_{outer}}{\partial \phi}}_{\textbf{direct grad.}} + \underbrace{\frac{\partial L_{outer}(\phi, \theta^\ast(\phi))}{\partial \theta^\ast(\phi)} \times \frac{\partial \theta^\ast(\phi)}{\partial{\phi}}}_{\textbf{indirect grad.}}.
\end{aligned}
\end{equation}
In our case, the \textbf{direct grad} is zero, thus we only need to calculate the \textbf{indirect grad}. However, the indirect grad is hard to compute as the best-response Jacobian term $\frac{\partial \theta^\ast(\phi)}{\partial{\phi}}$ needs the computation through the conditional training trajectories of $\min \theta^\ast(\phi)$ given $\phi$~\cite{lorraine2020optimizing}. One can use IFT to approximate the best-response Jacobian as follows 
\begin{equation}
\label{eq:indirect_gradient}
    \frac{\partial \theta^\ast(\phi)}{\partial \phi}=-\Big[\frac{\partial ^2 L_{inner}}{\partial \theta \partial \theta^{\rm T}}\Big]^{-1} \times \frac{\partial^2L_{inner}}{\partial \theta \partial \phi^{\rm T}}.
\end{equation}

\paragraph{Neumann series based IFT.} Inverting a Hessian matrix in high dimension is intractable. In our algorithm, we employ the Neumann series to approximate the Hessian inversion as per~\cite{lorraine2020optimizing}. The formulation is as follows
\begin{equation}
\label{eq:neumann}
    \Big[\frac{\partial ^2 L_{inner}}{\partial \theta \partial \theta^{\rm T}}\Big]^{-1} = \lim_{i\rightarrow\infty} \sum_{j=0}^{i}\Big[ I - \frac{\partial ^2 L_{inner}}{\partial \theta \partial \theta^{\rm T}}\Big].
\end{equation}
Thanks to the efficiency brought by this approximation, the augmentation module $\mathcal{A}_{\phi}$ and the model $F_{\theta}$ can be optimized efficiently during training. The pipeline is summarized as Algorithm \ref{alg:AugLearn}.

\begin{algorithm}[h]
\caption{Learning to Augment for DG}
\label{alg:AugLearn}
\SetAlgoLined
\textbf{Input}: Domain $\mathcal{D}$ \\
\textbf{Init}: Classification model parameters $\theta$. Augmentation module parameters $\phi$. Hyperparameters $\alpha, \beta$ \\
\For{ite \textbf{in} iterations}
{
        \textbf{Split}: $\hat{\mathcal{D}^{psrc}}$, $\hat{\mathcal{D}^{ptrg}}$ $\gets$ $\mathcal{D}$ \\
        \For{ite \textbf{in} inner iterations}
        {
          \textbf{Inner-loop}: Compute \text{gradients} $\frac{\partial L_{inner}}{\partial \theta}$ \\
          \textbf{Inner-loop optimization}: Update parameters  $\theta = \theta - \alpha \frac{\partial L_{inner}}{\partial \theta}$ \\
        }
        \textbf{Outer-loop}: Compute gradients $\frac{\partial L_{outer}(\phi)}{\partial \phi}$ according to Eq.~$5$, Eq.~$6$ and Eq.~$7$ in the main body.\\
       \textbf{Outer-loop optimization}: Updated parameters $\phi = \phi - \beta \frac{\partial L_{outer}}{\partial \phi}$ \\
}
\textbf{Output}: $\theta$

\end{algorithm}

\paragraph{Augmentation in frequency space.}
Our AugLearn is also capable of augmenting the input data in frequency space. To this end, we first convert the image into the frequency space using DCT, and then apply the augmentation module to the frequency spectrum. The inner loop optimization equation \ref{eq:inner} is then reformulated as
\begin{equation}
\theta^*= \underset{\theta}{\arg\min} \frac{1}{|\hat{\mathcal{D}^{src}}|} \sum_{x, y \sim \hat{\mathcal{D}^{src}}} \ell_{ce}(F_\theta(\mathcal{T}_{inv}(\mathcal{A}_{\phi}(\mathcal{T}(x)))), y),
\end{equation}
where $\mathcal{T}(\cdot)$ is DCT, and $\mathcal{T}_{inv}(\cdot)$ is inverse DCT. This variant is named as AugLearn-F. The overall training and inference of AugLearn-F are the same as the vanilla AugLearn method.

\subsection{Inference}
At model inference, the augmentation module is disabled. Given an input from the unseen domain $D_{S+1}$, the prediction is computed as
\begin{equation}
 \hat{y} = F_\theta(x), x \sim {D}_{S+1}.
\end{equation}
As $F_\theta(x)$ is trained with an augmentation module, which is optimized to generate augmented images against domain shift, the precision of $\hat{y}$ from our model is thus guaranteed.

\section{Experiments}
\subsection{Experimental Setup}
\paragraph{Datasets and settings.}
We evaluate our approach on three commonly used DG benchmark datasets, namely PACS~\cite{li2017deeper}, Office-Home~\cite{venkateswara2017deep} and Digits-DG~\cite{zhou2020deep}. (1) PACS is composed of four domains, which are art, cartoon, photo and sketch, with 9,991 images in total. There are seven classes for each one of these domains, and the domain shift mainly comes from dramatic style changes. 
(2) Office-Home consists of four domains including artistic, clipart, product and real-world. There are 65 object classes and approximately 15,500 images in total. Images from different domains differ in viewpoint, background and image style. (3) Digits-DG includes MNIST, MNIST-M, SVHN and SYN, which differ in font style, stroke and background. There are ten classes in each domain, and each class has $600$ images. We report the top-1 classification accuracy averaged over five runs with different random seeds. Due to the space constraint, the Digits-DG results are available in supplementary material.

\paragraph{Baselines.} We compare our method with the current state of the art DG methods including CCSA~\cite{motiian2017unified}, MMD-AAE~\cite{li2018adversarial}, CrossGrad~\cite{shankar2018generalizing}, JiGen~\cite{carlucci2019domain}, DDAIG~\cite{zhou2020deep}, L2A-OT~\cite{zhou2020learning} and MixStyle~\cite{zhou2021domain}. CCSA and MMD-AAE align the features from different source domains to a unified space and train a model based on those aligned features. CrossGrad is based on the domain classification guided image augmentation. JiGen introduces jigsaw solving to DG as an auxiliary task. DDAIG and L2A-OT are designed to generate novel domain images to improve the model generalization ability. MixStyle diversifies inputs by mixing the styles in feature space. We also compare with ERM, which serves as a strong baseline for DG. 

\subsection{Evaluation on PACS}
\paragraph{Implementation.} Following ~\cite{carlucci2019domain}, we use the ImageNet-pretrained ResNet$18$ as the feature extractor with a followed softmax classifier. %We parameterize the augmentation module as U-Net ~\cite{ronneberger2015unet}.%
All the images are resized to $224 \times 224$. The networks are trained with SGD, with initial learning rate of $1e-3$, batch size of $16$ and weight decay of $5e-4$ for $50$ epochs. The learning rate is decayed by $0.1$ at the $30$th epoch and the $20$th epoch for classification optimizer and augment-optimizer, respectively. 

\paragraph{Results.} 
The experimental results on PACS is shown in Table~\ref{tab:results} (left). First, we observe that our proposed augmentation modules AugLearn and AugLearn-F outperform the ERM baseline with remarkable accuracy margins $4.6\%$ and $4.8\%$ respectively, demonstrating the efficacy of our proposed algorithms.
%Meanwhile, our AugLearn outperforms the previous state of the art DG method MixStyle with $0.4\%$ accuracy improvement. 
More interestingly, our AugLearn variants are complementary with a feature augmentation based DG method, MixStyle. Incorporating our AugLearn(-F), it improves $1.3\%$ ($1.2\%$) accuracy over the vanilla MixStyle. Our proposed augmentation algorithm differs from DDAIG and L2A-OT regarding optimization objectives and training strategy. The results show that optimizing the augmentation module explicitly with the simulated domain shift can be much more effective than optimizing it using the complicated learning objectives on the source domain data. Our AugLearn(-F) outperforms all the other methods on cartoon and sketch domains, which are two most challenging held out domains, especially with a noticeable improvement ($5.6 \%$) on sketch. These results explain our assumption, that a pure image augmentation may be beneficial to model performance but is not robust to large domain shifts.

\begin{table*}[htbp]\Huge
%	\tabstyle{8pt}
	\centering
	\resizebox{1.0\linewidth}{!}{
	\begin{tabular}{l|cccc|c|cccc|c}
		\toprule
		\multirow{2}{*}{Method} & \multicolumn{5}{c|}{PACS} & \multicolumn{5}{c}{Office-Home} \\
		\cmidrule{2-11}
		    &  Art & Cartoon & Photo & Sketch & Average & Artistic & Clipart & Product & Real World & Average \\
		\midrule
		ERM       & 78.5 & 75.2 & 96.2 & 67.9 & 79.5 & 58.4 & 49.2 & 74.1 & 76.3 & 64.5 \\
		CCSA      & 80.5 & 76.9 & 93.6 & 66.8 & 79.4 & 59.9 & 49.9 & 74.1 & 75.7 & 64.9 \\
		MMD-AAE   & 75.2 & 72.7 & 96.0 & 64.2 & 77.0 & 56.5 & 47.3 & 72.1 & 74.8 & 62.7 \\
		CrossGrad & 79.8 & 76.8 & 96.0 & 70.2 & 80.7 & 58.4 & 49.4 & 73.9 & 75.8 & 64.4 \\
		JiGen     & 79.4 & 75.3 & 96.0 & 71.6 & 80.5 & 53.0 & 47.5 & 71.5 & 72.8 & 61.2 \\
		DDAIG     & \textbf{84.2} & 77.0 & 95.3 & 83.1 & 74.7 & 59.2 & 52.3 & 74.6 & 76.0 & 65.5 \\
		L2A-OT    & 83.3 & 78.2 & \textbf{96.2} & 82.8 & 73.6 & \textbf{60.6} & 50.1 & \textbf{74.8} & \textbf{77.0} & 65.6 \\
		MixStyle  & 84.1 & 78.8 & 96.1 & 75.9 & 83.7 & 58.7 & 53.4 & 74.2 & 75.9 & 65.5 \\
        \midrule
        ERM+AugLearn   & 82.9 & 78.8 & 94.5 & 80.1 & 84.1 (+4.6) & 58.9 & 53.3 & 74.3 & 76.0 & 65.6 (+1.1)\\
        ERM+AugLearn-F & 81.9 & \textbf{79.2} & 95.3 & 80.7 & 84.3 (+4.8) & 58.5 & \textbf{54.2} & 73.2 & 75.1 & 65.3 (+0.8)\\
        MixStyle+AugLearn & 84.1 & 79.0 & 95.2 & \textbf{81.5} & \textbf{85.0} (+1.3) & 59.3 & 53.5 & 74.6 & 76.0 & \textbf{66.0} (+0.5) \\
 		MixStyle+AugLearn-F & 83.9 & \textbf{79.2} & 95.4 & 81.0 & 84.9 (+1.2) & 59.8 & 52.7 & \textbf{74.8} & 75.6 & 65.7 (+0.2) \\
        
	\bottomrule
	\end{tabular}
	}
	\caption{Leave-one-domain-out generalization results on PACS and Office-Home.}
	\label{tab:results}
\end{table*}

\subsection{Evaluation on Office-Home}
\paragraph{Implementation.}
We use the training, validation splits following ~\cite{zhou2021domain}. We train the model on the training set from the source domains and test the trained model on the held out test domain. The training details are the same as those of PACS dataset.

\paragraph{Results.}
As shown in Table~\ref{tab:results} (right), our proposed method achieves again the state of the art performance on this benchmark, further demonstrating the effectiveness of our proposed AugLearn. Specifically, our AugLearn(-F) improves over the ERM baseline with a $1.1\%$ ($0.8\%$) accuracy margin. Meanwhile, our method achieves comparable results with L2A-OT, the recent augmentation based DG method. Again on the most challenging held out domain clipart, we achieve the best performance over all other competitors, demonstrating the robustness of our AugLearn(-F) against large domain shift. Our AugLearn(-F) still complements with MixStyle enabling a $0.5\%$ ($0.2\%$) accuracy gain over vanilla MixStyle.

% \textcolor{red}{From Table \ref{tab:results}, our proposed methods again achieve the state of the art performance to the previous DG methods and improve over ERM with $0.4\%$ and $1.5\%$ for AugLearn and AugLearn-F respectively. AugLearn grants strong generalization ability by diversifying the inputs with meta-learning. Another observation is still consistent that our methods improve MixStyle with $0.7\%$ and $2.1\%$, demonstrating the complementarity of these two kinds of augmentation methods which work in time and frequency spaces respectively. It is worth noting that MixStyle$+$AugLearn-F outperforms all the domains except for the SYN domain. The AugLearn-F and MixStyle$+$AugLearn-F perform better than the vanilla AugLearn and MixStyle$+$AugLearn on Digits-DG, demonstrating that AugLearn is capable of capturing style information and diversifying it more effectively in frequency space.}

\section{Further Analysis}
\subsection{Ablation Study}
We conduct further experiments to analyze our proposed AugLearn(-F). We attribute the performance of our AugLearn to the simulated domain shift and meta learning. We thus compare it with a simple variant (w/o meta learning), which updates the classification model and the augmentation module on the pseudo source and pseudo target domains respectively, without bilevel optimization. From the results in Table~\ref{tab:ablation}, we observe that optimizing the augmentation module during training is indeed helpful with $3.2\%$ and $3.5\%$ accuracy improvements over the ERM baseline with regard to AugLearn and AugLearn-F, respectively. Adding meta learning with domain shift improves the model performance further by $1.4\%$ and $1.3\%$ accuracy margins. These results demonstrate the efficacy of our proposed meta learning pipeline.

\subsection{Visualization of Augmented Images}
Figure~\ref{fig:visualization} illustrates the outputs of the augmentation module trained on two used DG benchmarks. We can see that the generated images (middle) by AugLearn are different from the original images (left) dramatically. The augmentation module not only changes the background but also diversifies the appearance of objects. These illustrative results show that our AugLearn indeed generates augmented images which are different from the vanilla ones but with the same semantic meaning without using explicit learning objectives.  We attribute this to our meta learning pipeline, which enables easy training. \textcolor{black}{Another interesting observation is that the images generated by AugLearn-F do not change much from the original images, i.e. Figure~\ref{fig:visualization} left v.s. right, while AugLearn-F still can improve the model performance clearly. This shows it is more effective to augment images in frequency space than that in time space.}

\subsection{Side Benefit of Adversarial Defence}
We also include the investigation of our models against adversarial attacks, such as FGSM ~\cite{goodfellow2014explaining}. After training the DG models, i.e. ERM, ERM with strong augmentations (Cutout~\cite{devries2017improved}, CutMix~\cite{yun2019cutmix} and DropBlock~\cite{ghiasi2018dropblock}) and AugLearn variants, we attack the trained DG model using FGSM attack with different strengths. From the results in Figure~\ref{fig:robustness}, we can see that after incorporating our AugLearn modules the attack success rate decreases significantly though the models are trained on clean data only. AugLearn(-F) enables the model to be more robust against FGSM attack while Cutout and Cutmix fail. Interestingly, we can see that injecting the AugLearn-F module during the training brings more model robustness than AugLearn against FGSM attack. It is found that the adversarial attacks conduct more perturbations in the middle and high frequencies of the vanilla image~\cite{freqrobustcnn2020}. Therefore, augmenting the input images in the frequency space during training may give more robustness to the trained model against potential adversarial attacks than augmenting in the time space.

\begin{minipage}{0.35\textwidth}
\centering
\includegraphics[width=1.0\textwidth]{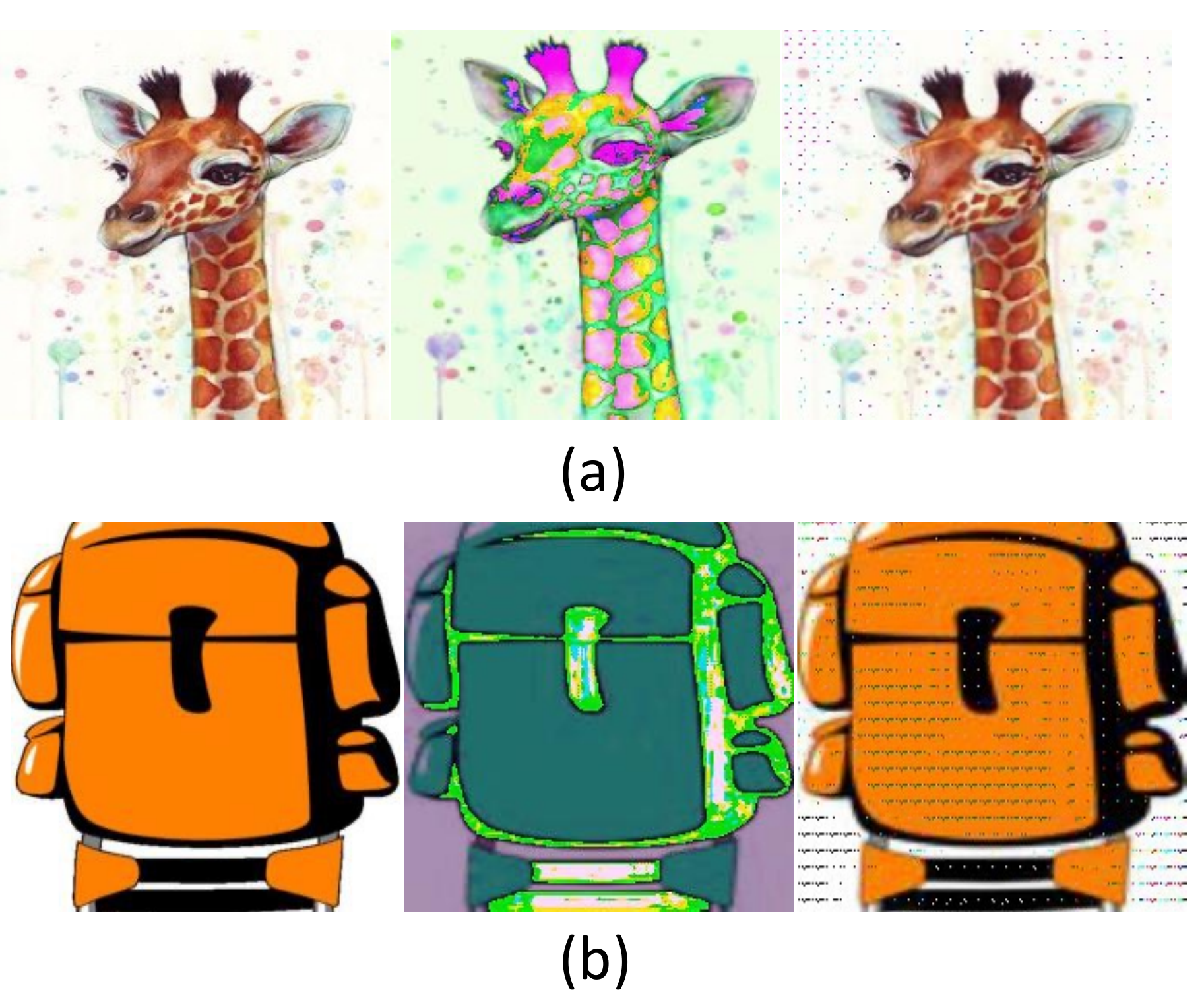}
\vspace*{2mm}
\captionof{figure}{Synthesized examples on PACS (a) and Office-Home (b). Left: raw, middle: AugLearn generated, right: AugLearn-F generated.}
\label{fig:visualization}
\end{minipage}
%   ~\hfill~
\begin{minipage}{0.65\textwidth}
\centering
\resizebox{0.9\linewidth}{!}{
    % \footnotesize
	\begin{tabular}{l|cccc|c}
		\toprule
		Method    &  Art & Cartoon & Photo & Sketch & Avg. \\
		\midrule
		ERM   & 78.5 & 75.2 & 96.2 & 67.9 & 79.5 \\
        AugLearn & 82.9 & 78.8 & 94.5 & 80.1 & 84.1 \\
 		\quad - w/o ML & 81.6 & 76.3 & 93.8 & 79.0 & 82.7 \\
 		AugLearn-F & 81.9 & 79.2 & 95.3 & 80.7 & 84.3 \\
 		\quad - w/o ML & 81.2 & 76.5 & 94.2 & 79.9 & 83.0 \\
		\bottomrule
	\end{tabular}
	}
	\vspace*{3mm}
    \captionof{table}{Ablation study results on PACS}
    \label{tab:ablation}
    \vspace{1mm}
    \includegraphics[width=0.85\textwidth]{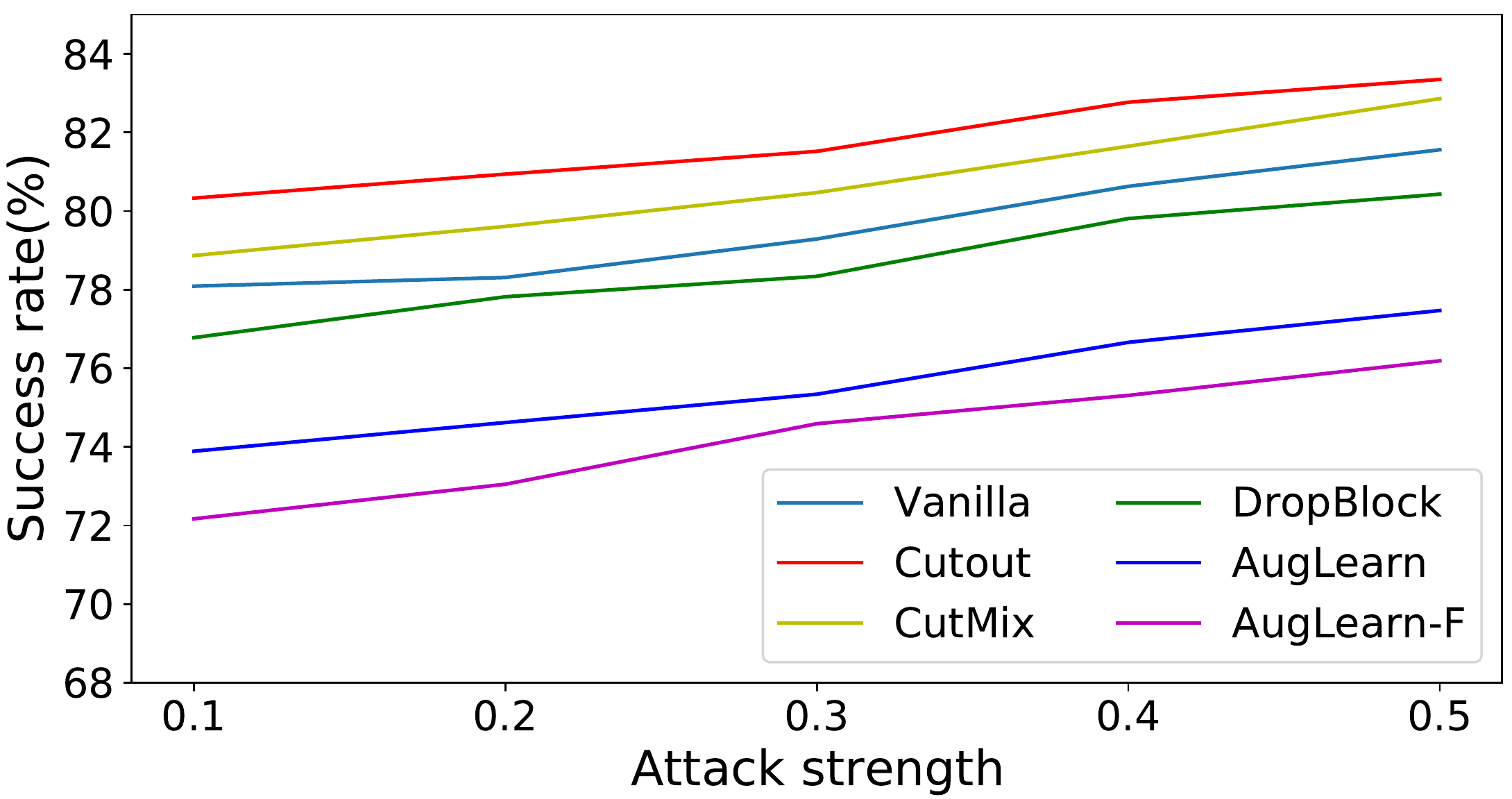}
    \vspace*{2.5mm}
    \captionof{figure}{\textcolor{black}{Success rate of FGSM attack on PACS.}}
    \label{fig:robustness}
\end{minipage}

\begin{wraptable}{r}{5cm}\Huge
% 	\tabstyle{8pt}
	\centering
	\resizebox{1.0\linewidth}{!}{
	\begin{tabular}{c|cccc|c}
		\toprule
		Method    &  backbone & extra parameters \\
		\midrule
		CrossGrad & \multirow{4}*{11.18m} & 11.17m \\
		DDAIG     &  & 0.23m \\
		L2A-OT    &  & 3.70m \\
		AugLearn  &  & 6.65k \\
		\bottomrule
	\end{tabular}}
	\caption{Number of trainable parameters of different methods.}
	\label{tab:NumParameters}
\end{wraptable}

\section{Compact Augmentation Module}
We also analyse the extra parameters introduced in the model training of different augmentation based DG methods. 
%The existing augmentation based DG methods typically require complicated training objectives, diversity loss, fidelity loss and semantic loss~\cite{zhou2020learning}.
From the numbers in Table~\ref{tab:NumParameters}, we can see that our AugLearn introduces extremely few trainable parameters compared with CrossGrad, DDAIG and L2A-OT. Specifically, we can see that CrossGrad doubles the total training parameters by introducing the augmentation module. Then, the method with the second most extra parameters is L2A-OT, which adds $3.70$m parameters. The tier three is DDAIG with $0.23$m extra parameters. Nevertheless, they brought many orders of magnitude more parameters compared to our AugLearn.
We assume this is due to the nature of our proposed meta learning pipeline, such that our augmentation module is effective though trained with much fewer parameters.

\section{Conclusion}
\textcolor{black}{We have presented a novel data augmentation based DG method, termed AugLearn. AugLearn treats the augmentation module as the model hyperparameters and optimizes it with meta learning.} Our AugLearn is light-weight, model-agnostic and applicable to any base DG methods (verified with two different DG methods). More inherently, our AugLearn module is capable of augmenting data in both the time and frequency spaces. Extensive experiments demonstrate that AugLearn variants achieve the state of the art performance on two popular DG benchmarks. Qualitative visualizations further explain that our AugLearn is able to generate augmented images which are different from the vanilla images in terms of both the foreground and background, with a simple meta learning objective.

\bibliography{egbib}

\begin{thebibliography}{55}
\providecommand{\natexlab}[1]{#1}
\providecommand{\url}[1]{\texttt{#1}}
\expandafter\ifx\csname urlstyle\endcsname\relax
  \providecommand{\doi}[1]{doi: #1}\else
  \providecommand{\doi}{doi: \begingroup \urlstyle{rm}\Url}\fi

\bibitem[Baktashmotlagh et~al.(2013)Baktashmotlagh, Harandi, Lovell, and
  Salzmann]{baktashmotlagh2013unsupervised}
M.~Baktashmotlagh, M.~T. Harandi, B.~C. Lovell, and M.~Salzmann.
\newblock Unsupervised domain adaptation by domain invariant projection.
\newblock In \emph{Proceedings of the IEEE International Conference on Computer
  Vision}, pages 769--776, 2013.

\bibitem[Balaji et~al.(2018)Balaji, Sankaranarayanan, and
  Chellappa]{balaji2018metareg}
Y.~Balaji, S.~Sankaranarayanan, and R.~Chellappa.
\newblock Metareg: Towards domain generalization using meta-regularization.
\newblock In \emph{Advances in Neural Information Processing Systems}, pages
  998--1008, 2018.

\bibitem[Ben-David et~al.(2010)Ben-David, Blitzer, Crammer, Kulesza, Pereira,
  and Vaughan]{ben2010theory}
S.~Ben-David, J.~Blitzer, K.~Crammer, A.~Kulesza, F.~Pereira, and J.~W.n
  Vaughan.
\newblock A theory of learning from different domains.
\newblock \emph{Machine learning}, 79\penalty0 (1):\penalty0 151--175, 2010.

\bibitem[Bengio(2000)]{bengio2000gradient}
Y.~Bengio.
\newblock Gradient-based optimization of hyperparameters.
\newblock \emph{Neural computation}, 12\penalty0 (8):\penalty0 1889--1900,
  2000.

\bibitem[Blanchard et~al.(2011)Blanchard, Lee, and
  Scott]{blanchard2011generalizing}
G.~Blanchard, G.~Lee, and C.~Scott.
\newblock Generalizing from several related classification tasks to a new
  unlabeled sample.
\newblock volume~24, pages 2178--2186, 2011.

\bibitem[Carlucci et~al.(2019)Carlucci, D'Innocente, Bucci, and
  Caputo]{carlucci2019domain}
F.~M. Carlucci, A.~D'Innocente, S.~Bucci, and T.~Caputo, B.and~Tommasi.
\newblock Domain generalization by solving jigsaw puzzles.
\newblock In \emph{Proceedings of the IEEE Conference on Computer Vision and
  Pattern Recognition}, pages 2229--2238, 2019.

\bibitem[DeVries and Taylor(2017)]{devries2017improved}
T.~DeVries and G.~W. Taylor.
\newblock Improved regularization of convolutional neural networks with cutout.
\newblock \emph{arXiv preprint arXiv:1708.04552}, 2017.

\bibitem[Ding and Fu(2017)]{ding2017deep}
Z.~Ding and Y.~Fu.
\newblock Deep domain generalization with structured low-rank constraint.
\newblock \emph{IEEE Transactions on Image Processing}, 27\penalty0
  (1):\penalty0 304--313, 2017.

\bibitem[Dou et~al.(2019)Dou, Castro, Kamnitsas, and Glocker]{dou2019domain}
Q.~Dou, D.~C. Castro, K.~Kamnitsas, and B.~Glocker.
\newblock Domain generalization via model-agnostic learning of semantic
  features.
\newblock In \emph{Advances in Neural Information Processing Systems}, pages
  6450--6461, 2019.

\bibitem[Du et~al.(2020)Du, Zhen, Shao, and Snoek]{du2020metanorm}
Y.~Du, X.~Zhen, L.~Shao, and C.~G. Snoek.
\newblock Metanorm: Learning to normalize few-shot batches across domains.
\newblock In \emph{International Conference on Learning Representations}, 2020.

\bibitem[Erfani et~al.(2016)Erfani, Baktashmotlagh, Moshtaghi, Nguyen, Leckie,
  Bailey, and Kotagiri]{erfani2016robust}
S.~Erfani, M.~Baktashmotlagh, M.~Moshtaghi, X.~Nguyen, C.~Leckie, J.~Bailey,
  and R.~Kotagiri.
\newblock Robust domain generalisation by enforcing distribution invariance.
\newblock In \emph{Proceedings of the Twenty-Fifth International Joint
  Conference on Artificial Intelligence (IJCAI-16)}, pages 1455--1461. AAAI
  Press, 2016.

\bibitem[Finn et~al.(2017)Finn, Abbeel, and Levine]{finn2017model}
C.~Finn, P.~Abbeel, and S.~Levine.
\newblock Model-agnostic meta-learning for fast adaptation of deep networks.
\newblock In \emph{International Conference on Machine Learning}, pages
  1126--1135. PMLR, 2017.

\bibitem[Ganin and Lempitsky(2015)]{ganin2015unsupervised}
Y.~Ganin and V.~Lempitsky.
\newblock Unsupervised domain adaptation by backpropagation.
\newblock In \emph{International conference on machine learning}, pages
  1180--1189. PMLR, 2015.

\bibitem[Ghiasi et~al.(2018)Ghiasi, Lin, and Le]{ghiasi2018dropblock}
G.~Ghiasi, T.~Lin, and Q.~V. Le.
\newblock Dropblock: A regularization method for convolutional networks.
\newblock \emph{arXiv preprint arXiv:1810.12890}, 2018.

\bibitem[Ghifary et~al.(2016)Ghifary, Balduzzi, Kleijn, and
  Zhang]{ghifary2016scatter}
M.~Ghifary, D.~Balduzzi, W.~B. Kleijn, and M.~Zhang.
\newblock Scatter component analysis: A unified framework for domain adaptation
  and domain generalization.
\newblock \emph{IEEE transactions on pattern analysis and machine
  intelligence}, 39\penalty0 (7):\penalty0 1414--1430, 2016.

\bibitem[Gong et~al.(2012)Gong, Shi, Sha, and Grauman]{gong2012geodesic}
B.~Gong, Y.~Shi, F.~Sha, and K.~Grauman.
\newblock Geodesic flow kernel for unsupervised domain adaptation.
\newblock In \emph{2012 IEEE conference on computer vision and pattern
  recognition}, pages 2066--2073. IEEE, 2012.

\bibitem[Goodfellow et~al.(2015)Goodfellow, Shlens, and
  Szegedy]{goodfellow2014explaining}
I.~J. Goodfellow, J.~Shlens, and C.~Szegedy.
\newblock Explaining and harnessing adversarial examples.
\newblock 2015.

\bibitem[Huang et~al.(2020)Huang, Wang, Xing, and Huang]{huang2020self}
Z.~Huang, H.~Wang, E.~P. Xing, and D.~Huang.
\newblock Self-challenging improves cross-domain generalization.
\newblock In \emph{Computer Vision--ECCV 2020: 16th European Conference,
  Glasgow, UK, August 23--28, 2020, Proceedings, Part II 16}, pages 124--140.
  Springer, 2020.

\bibitem[Jeon et~al.(2021)Jeon, Hong, Lee, Lee, and Byun]{jeon2021feature}
S.~Jeon, K.~Hong, P.~Lee, J.~Lee, and H.~Byun.
\newblock Feature stylization and domain-aware contrastive learning for domain
  generalization.
\newblock In \emph{Proceedings of the 29th ACM International Conference on
  Multimedia}, pages 22--31, 2021.

\bibitem[Jin et~al.(2020)Jin, Lan, Zeng, and Chen]{jin2020feature}
X.~Jin, C.~Lan, W.~Zeng, and Z.~Chen.
\newblock Feature alignment and restoration for domain generalization and
  adaptation.
\newblock \emph{arXiv preprint arXiv:2006.12009}, 2020.

\bibitem[Krizhevsky et~al.(2012)Krizhevsky, Sutskever, and
  Hinton]{krizhevsky2012imagenet}
A.~Krizhevsky, I.~Sutskever, and G.~E. Hinton.
\newblock Imagenet classification with deep convolutional neural networks.
\newblock volume~25, pages 1097--1105, 2012.

\bibitem[Li et~al.(2017)Li, Yang, Song, and Hospedales]{li2017deeper}
D.~Li, Y.~Yang, Y.~Song, and T.~M. Hospedales.
\newblock Deeper, broader and artier domain generalization.
\newblock In \emph{Proceedings of the IEEE international conference on computer
  vision}, pages 5542--5550, 2017.

\bibitem[Li et~al.(2018{\natexlab{a}})Li, Yang, Song, and
  Hospedales]{li2018learning}
D.~Li, Y.~Yang, Y.~Song, and T.~M. Hospedales.
\newblock Learning to generalize: Meta-learning for domain generalization.
\newblock In \emph{Thirty-Second AAAI Conference on Artificial Intelligence},
  2018{\natexlab{a}}.

\bibitem[Li et~al.(2019{\natexlab{a}})Li, Zhang, Yang, Liu, Song, and
  Hospedales]{li2019episodic}
D.~Li, J.~Zhang, Y.~Yang, C.~Liu, Y.~Song, and T.~M. Hospedales.
\newblock Episodic training for domain generalization.
\newblock In \emph{Proceedings of the IEEE International Conference on Computer
  Vision}, pages 1446--1455, 2019{\natexlab{a}}.

\bibitem[Li et~al.(2018{\natexlab{b}})Li, Pan, Wang, and
  Kot]{li2018adversarial}
H.~Li, J.~S. Pan, S.~Wang, and A.~C. Kot.
\newblock Domain generalization with adversarial feature learning.
\newblock In \emph{Proceedings of the IEEE Conference on Computer Vision and
  Pattern Recognition}, pages 5400--5409, 2018{\natexlab{b}}.

\bibitem[Li et~al.(2021)Li, Li, Li, Gong, Fu, and Hospedales]{li2021simple}
P.~Li, D.~Li, W.~Li, S.~Gong, Y.~Fu, and T.~M. Hospedales.
\newblock A simple feature augmentation for domain generalization.
\newblock In \emph{Proceedings of the IEEE/CVF International Conference on
  Computer Vision}, pages 8886--8895, 2021.

\bibitem[Li et~al.(2019{\natexlab{b}})Li, Yang, Zhou, and
  Hospedales]{li2019feature}
Y.~Li, Y.~Yang, W.~Zhou, and T.~Hospedales.
\newblock Feature-critic networks for heterogeneous domain generalization.
\newblock In \emph{International Conference on Machine Learning}, pages
  3915--3924. PMLR, 2019{\natexlab{b}}.

\bibitem[Long et~al.(2014)Long, Wang, Ding, Sun, and Yu]{long2014transfer}
M.~Long, J.~Wang, G.~Ding, J.~Sun, and P.~S. Yu.
\newblock Transfer joint matching for unsupervised domain adaptation.
\newblock In \emph{Proceedings of the IEEE conference on computer vision and
  pattern recognition}, pages 1410--1417, 2014.

\bibitem[Lorraine et~al.(2020)Lorraine, Vicol, and
  Duvenaud]{lorraine2020optimizing}
J.~Lorraine, P.~Vicol, and D.~Duvenaud.
\newblock Optimizing millions of hyperparameters by implicit differentiation.
\newblock In \emph{International Conference on Artificial Intelligence and
  Statistics}, pages 1540--1552. PMLR, 2020.

\bibitem[Luketina et~al.(2016)Luketina, Berglund, Greff, and
  Raiko]{luketina2016scalable}
J.~Luketina, M.~Berglund, K.~Greff, and T.~Raiko.
\newblock Scalable gradient-based tuning of continuous regularization
  hyperparameters.
\newblock In \emph{International conference on machine learning}, pages
  2952--2960. PMLR, 2016.

\bibitem[Matsuura and Harada(2020)]{matsuura2020domain}
T.~Matsuura and T.~Harada.
\newblock Domain generalization using a mixture of multiple latent domains.
\newblock In \emph{Proceedings of the AAAI Conference on Artificial
  Intelligence}, volume~34, pages 11749--11756, 2020.

\bibitem[Moreno-Torres et~al.(2012)Moreno-Torres, Raeder, Alaiz-Rodr{\'\i}guez,
  Chawla, and Herrera]{moreno2012unifying}
J.~G. Moreno-Torres, T.~Raeder, R.~Alaiz-Rodr{\'\i}guez, N.~V. Chawla, and
  F.~Herrera.
\newblock A unifying view on dataset shift in classification.
\newblock \emph{Pattern recognition}, 45\penalty0 (1):\penalty0 521--530, 2012.

\bibitem[Motiian et~al.(2017)Motiian, Piccirilli, Adjeroh, and
  Doretto]{motiian2017unified}
S.~Motiian, M.~Piccirilli, D.~A. Adjeroh, and G.~Doretto.
\newblock Unified deep supervised domain adaptation and generalization.
\newblock In \emph{Proceedings of the IEEE International Conference on Computer
  Vision}, pages 5715--5725, 2017.

\bibitem[Muandet et~al.(2013)Muandet, Balduzzi, and
  Sch{\"o}lkopf]{muandet2013domain}
K.~Muandet, D.~Balduzzi, and B.~Sch{\"o}lkopf.
\newblock Domain generalization via invariant feature representation.
\newblock In \emph{International Conference on Machine Learning}, pages 10--18,
  2013.

\bibitem[Ot{\'a}lora et~al.(2019)Ot{\'a}lora, Atzori, Andrearczyk, Khan, and
  M{\"u}ller]{otalora2019staining}
Se. Ot{\'a}lora, M.~Atzori, V.~Andrearczyk, A.~Khan, and H.~M{\"u}ller.
\newblock Staining invariant features for improving generalization of deep
  convolutional neural networks in computational pathology.
\newblock \emph{Frontiers in bioengineering and biotechnology}, 7:\penalty0
  198, 2019.

\bibitem[Rajeswaran et~al.(2019)Rajeswaran, Finn, Kakade, and
  Levine]{rajeswaran2019meta}
A.~Rajeswaran, C.~Finn, S.~Kakade, and S.~Levine.
\newblock Meta-learning with implicit gradients.
\newblock \emph{Advances in neural information processing systems}, 2019.

\bibitem[Ravi and Larochelle(2017)]{ravi2016optimization}
S.~Ravi and H.~Larochelle.
\newblock Optimization as a model for few-shot learning.
\newblock In \emph{ICLR}, 2017.

\bibitem[Recht et~al.(2019)Recht, Roelofs, Schmidt, and
  Shankar]{recht2019imagenet}
B.~Recht, R.~Roelofs, L.~Schmidt, and V.~Shankar.
\newblock Do imagenet classifiers generalize to imagenet?
\newblock In \emph{International Conference on Machine Learning}, pages
  5389--5400. PMLR, 2019.

\bibitem[Ronneberger et~al.(2015)Ronneberger, Fischer, and
  Brox]{ronneberger2015unet}
O.~Ronneberger, P.~Fischer, and T.~Brox.
\newblock U-net: Convolutional networks for biomedical image segmentation.
\newblock In \emph{International Conference on Medical image computing and
  computer-assisted intervention}, pages 234--241. Springer, 2015.

\bibitem[Shankar et~al.(2018)Shankar, Piratla, Chakrabarti, Chaudhuri, Jyothi,
  and Sarawagi]{shankar2018generalizing}
S.~Shankar, V.~Piratla, S.~Chakrabarti, S.~Chaudhuri, P.~Jyothi, and
  S.~Sarawagi.
\newblock Generalizing across domains via cross-gradient training.
\newblock In \emph{Proceedings of the ICLR}, 2018.

\bibitem[Taori et~al.(2012)Taori, Dave, Shankar, Carlini, Recht, and
  Schmidt]{taori2020measuring}
R.~Taori, A.~Dave, V.~Shankar, N.~Carlini, B.~Recht, and L.~Schmidt.
\newblock Measuring robustness to natural distribution shifts in image
  classification.
\newblock \emph{Advances in neural information processing systems}, 2012.

\bibitem[Venkateswara et~al.(2017)Venkateswara, Eusebio, Chakraborty, and
  Panchanathan]{venkateswara2017deep}
H.~Venkateswara, J.~Eusebio, S.~Chakraborty, and S.~Panchanathan.
\newblock Deep hashing network for unsupervised domain adaptation.
\newblock In \emph{Proceedings of the IEEE Conference on Computer Vision and
  Pattern Recognition}, pages 5018--5027, 2017.

\bibitem[Volpi and Murino(2019)]{volpi2019addressing}
R.~Volpi and V.~Murino.
\newblock Addressing model vulnerability to distributional shifts over image
  transformation sets.
\newblock In \emph{Proceedings of the IEEE/CVF International Conference on
  Computer Vision}, pages 7980--7989, 2019.

\bibitem[Wang et~al.(2021)Wang, Lan, Liu, Ouyang, Zeng, and
  Qin]{wang2021generalizing}
J.~Wang, C.~Lan, C.~Liu, Y.~Ouyang, W.~Zeng, and T.~Qin.
\newblock Generalizing to unseen domains: A survey on domain generalization.
\newblock \emph{arXiv preprint arXiv:2103.03097}, 2021.

\bibitem[Wang et~al.(2020)Wang, Yang, Shrivastava, Rawal, and
  Ding]{freqrobustcnn2020}
Z.~Wang, Y.~Yang, A.~Shrivastava, V.~Rawal, and Z.~Ding.
\newblock Towards frequency-based explanation for robust {CNN}.
\newblock \emph{CoRR}, abs/2005.03141, 2020.
\newblock URL \url{https://arxiv.org/abs/2005.03141}.

\bibitem[Xu et~al.(2021)Xu, Zhang, Zhang, Wang, and Tian]{xu2021fourier}
Q.~Xu, R.~Zhang, Y.~Zhang, Y.~Wang, and Q.~Tian.
\newblock A fourier-based framework for domain generalization.
\newblock In \emph{Proceedings of the IEEE/CVF Conference on Computer Vision
  and Pattern Recognition}, pages 14383--14392, 2021.

\bibitem[Yue et~al.(2019)Yue, Zhang, Zhao, Sangiovanni-Vincentelli, Keutzer,
  and Gong]{yue2019domain}
X.~Yue, Y.~Zhang, S.~Zhao, A.~Sangiovanni-Vincentelli, K.~Keutzer, and B.~Gong.
\newblock Domain randomization and pyramid consistency: Simulation-to-real
  generalization without accessing target domain data.
\newblock In \emph{Proceedings of the IEEE/CVF International Conference on
  Computer Vision}, pages 2100--2110, 2019.

\bibitem[Yun et~al.(2019)Yun, Han, Oh, Chun, Choe, and Yoo]{yun2019cutmix}
S.~Yun, D.~Han, S.~J. Oh, S.~Chun, J.~Choe, and Y.~Yoo.
\newblock Cutmix: Regularization strategy to train strong classifiers with
  localizable features.
\newblock In \emph{Proceedings of the IEEE/CVF International Conference on
  Computer Vision}, pages 6023--6032, 2019.

\bibitem[Zakharov et~al.(2019)Zakharov, Kehl, and
  Ilic]{zakharov2019deceptionnet}
S.~Zakharov, W.~Kehl, and S.~Ilic.
\newblock Deceptionnet: Network-driven domain randomization.
\newblock In \emph{Proceedings of the IEEE International Conference on Computer
  Vision}, pages 532--541, 2019.

\bibitem[Zhang et~al.(2017)Zhang, Cisse, Dauphin, and
  Lopez-Paz]{zhang2017mixup}
H.~Zhang, M.~Cisse, Y.~N. Dauphin, and D.~Lopez-Paz.
\newblock mixup: Beyond empirical risk minimization.
\newblock \emph{arXiv preprint arXiv:1710.09412}, 2017.

\bibitem[Zhang et~al.(2020)Zhang, Wang, Yang, Sanford, Harmon, Turkbey, Wood,
  Roth, Myronenko, Xu, et~al.]{zhang2020generalizing}
L.~Zhang, X.~Wang, D.~Yang, T.~Sanford, S.~Harmon, B.~Turkbey, B.~J. Wood,
  H.~Roth, A.~Myronenko, D.~Xu, et~al.
\newblock Generalizing deep learning for medical image segmentation to unseen
  domains via deep stacked transformation.
\newblock \emph{IEEE transactions on medical imaging}, 39\penalty0
  (7):\penalty0 2531--2540, 2020.

\bibitem[Zhou et~al.(2020{\natexlab{a}})Zhou, Yang, Hospedales, and
  Xiang]{zhou2020learning}
K.~Zhou, Y.~Yang, T.~Hospedales, and T.~Xiang.
\newblock Learning to generate novel domains for domain generalization.
\newblock In \emph{European Conference on Computer Vision}, pages 561--578.
  Springer, 2020{\natexlab{a}}.

\bibitem[Zhou et~al.(2020{\natexlab{b}})Zhou, Yang, Hospedales, and
  Xiang]{zhou2020deep}
K.~Zhou, Y.~Yang, T.~M. Hospedales, and T.~Xiang.
\newblock Deep domain-adversarial image generation for domain generalisation.
\newblock In \emph{AAAI}, pages 13025--13032, 2020{\natexlab{b}}.

\bibitem[Zhou et~al.(2021{\natexlab{a}})Zhou, Liu, Qiao, Xiang, and
  Loy]{zhou2021survey}
K.~Zhou, Z.~Liu, Y.~Qiao, T.~Xiang, and C.~C. Loy.
\newblock Domain generalization: A survey.
\newblock \emph{arXiv preprint arXiv:2103.02503}, 2021{\natexlab{a}}.

\bibitem[Zhou et~al.(2021{\natexlab{b}})Zhou, Yang, Qiao, and
  Xiang]{zhou2021domain}
K.~Zhou, Y.~Yang, Y.~Qiao, and T.~Xiang.
\newblock Domain generalization with mixstyle.
\newblock In \emph{Proceedings of the ICLR}, 2021{\natexlab{b}}.

\end{thebibliography}
\end{document}